\documentclass{article}

\usepackage{arxiv}

\usepackage[utf8]{inputenc} 
\usepackage[T1]{fontenc}    
\usepackage{hyperref}       
\usepackage{url}            
\usepackage{booktabs}       
\usepackage{amsfonts}       
\usepackage{nicefrac}       
\usepackage{microtype}      
\usepackage{lipsum}

\usepackage{amsmath,amsfonts,bm}

\def\eqref#1{equation~\ref{#1}}

\def\1{\bm{1}}

\def\ve{{\bm{e}}}

\def\vi{{\bm{i}}}

\def\vo{{\bm{o}}}
\def\vp{{\bm{p}}}

\def\mB{{\bm{B}}}

\def\mD{{\bm{D}}}

\def\mG{{\bm{G}}}

\def\mO{{\bm{O}}}

\def\mS{{\bm{S}}}

\def\mW{{\bm{W}}}
\def\mX{{\bm{X}}}

\DeclareMathAlphabet{\mathsfit}{\encodingdefault}{\sfdefault}{m}{sl}
\SetMathAlphabet{\mathsfit}{bold}{\encodingdefault}{\sfdefault}{bx}{n}

\usepackage{textcomp}
\usepackage{graphicx}
\usepackage{caption}
\usepackage{subcaption}
\usepackage{array}
\newcolumntype{C}[1]{>{\centering\arraybackslash}p{#1}}
\usepackage{multirow}
\usepackage{amsmath}
\usepackage[toc,page,titletoc]{appendix}

\newcommand\blfootnote[1]{%
  \begingroup
  \renewcommand\thefootnote{}\footnote{#1}%
  \addtocounter{footnote}{-1}%
  \endgroup
}
\usepackage{amssymb}
\usepackage[utf8]{inputenc}
\usepackage{geometry}
\usepackage{algorithm}
\usepackage{algpseudocode}

\title{Extension of Direct Feedback Alignment to Convolutional and Recurrent Neural Network for Bio-plausible Deep Learning}

\author{Donghyeon Han$^{1)}$, Gwangtae Park$^{1)}$, Junha Ryu, Hoi-jun Yoo \\
School of Electrical Engineering\\
Korea Advanced Institute of Science and Technology (KAIST)\\
Daejeon, Republic of Korea \\
\texttt{\{hdh4797,gwangtaepark,junha.ryu,hjyoo\}@kaist.ac.kr} \\
}

\begin{document}
\maketitle

\begin{abstract}\blfootnote{1) These authors contributed equally}
Throughout this paper, we focus on the improvement of the direct feedback alignment (DFA) algorithm and extend the usage of the DFA to convolutional and recurrent neural networks (CNNs and RNNs). Even though the DFA algorithm is biologically plausible and has a potential of high-speed training, it has not been considered as the substitute for back-propagation (BP) due to the low accuracy in the CNN and RNN training. In this work, we propose a new DFA algorithm for BP-level accurate CNN and RNN training. Firstly, we divide the network into several modules and apply the DFA algorithm within the module. Second, the DFA with the sparse backward weight is applied. It comes with a form of dilated convolution in the CNN case, and in a form of sparse matrix multiplication in the RNN case. Additionally, the error propagation method of CNN becomes simpler through the group convolution. Finally, hybrid DFA increases the accuracy of the CNN and RNN training to the BP-level while taking advantage of the parallelism and hardware efficiency of the DFA algorithm. 


\end{abstract}

\section{Introduction}
Recently, deep neural networks (DNN) have enabled various AI-driven tasks and the back-propagation (BP) algorithm is recognized as the typical training methodology of the DNNs. However, the BP algorithm is regarded to be biologically implausible for two reasons: (1) back-propagation enforces the backward weight to be always symmetric to the forward weight (weight transport problem, \cite{WT}), and (2) the error should be propagated layer-by-layer (backward locking problem, \cite{BL}). The weight transport problem induces additional memory transactions caused by the reading of transposed backward weight at each learning iteration. The backward locking problem makes it hard to parallelize the DNN training since the next feedforward step cannot proceed until the entire back-propagation step is finished.

Several biologically plausible learning algorithms have been proposed to overcome the limitation of the back-propagation. Target propagation (\cite{TP}) relieved the weight transport problem by using the formulated local loss function instead of the error gradient to update the intermediate layers. Feedback alignment suggested by \cite{FA} used the random and fixed backward weight for error propagation instead of transposed forward weight. \cite{DFA} solved both the weight transport and backward locking problem by using the direct feedback alignment (DFA). It directly propagates errors from the last output layer to the intermediate layers using the random backward weight. However, those biologically plausible learning algorithms failed to train the convolutional neural networks (CNNs) or recurrent neural networks (RNNs) due to the large accuracy drop.

\cite{BDFA} showed the CNN can be successfully trained on CIFAR-10 and CIFAR-100 dataset by adopting DFA to the FC layers but back-propagation to the convolutional layers. However, the backward locking problem was partially solved only in the FC layers, thereby it missed the opportunity of parallelism in the error propagation of convolutional layers. 

In this paper, we suggest new methodologies to improve the accuracy of the DFA algorithm in the CNN and RNN training. The main contributions of our work are summarized as follows. 
\begin{enumerate}
\item[$\bullet$] Firstly, we divide both the CNN and RNN into several small modules and enables direct error propagation inside the module.
\item[$\bullet$] Secondly, we modify the DFA algorithm to be applied to the CNN training by using the following features: 
    \begin{enumerate}
    \item Dilated-convolution based error propagation is adopted instead of a fully-connected manner.
    \item Error propagation is simplified with the group convolution suggested by \cite{ShuffleNet}.
    \end{enumerate}
\item[$\bullet$] Thirdly, we simplify the backward weight used in the RNN training by adopting the upper triangular matrix. Moreover, we combine conventional DFA optimization methods such as binary DFA (\cite{BDFA}) and sparse DFA (\cite{SDFA}).
\item[$\bullet$] Fourthly, we propose a new training algorithm, hybrid DFA (HDFA) which mixes both BP and DFA based error propagation to achieve BP-level accuracy with the minimum hardware cost. 
\item[$\bullet$] Finally, we verified the performance of the proposed method by using the three different network architectures in both CNN and RNN training. Experimental results successfully demonstrate that the proposed algorithm can achieve high training accuracy as same as BP. Moreover, it shows the lowest memory transaction amounts compared with the existing error propagation methodologies.  
\end{enumerate}

The remaining part of the paper is organized as follows. The mathematical notation of the BP and DFA will be explained and the problem of the BP based algorithm, backward locking, will be introduced in Section 2. Then, the detailed explanation of the proposed algorithm and the related experiments will be followed in Section 3 and 4. The paper will be concluded in Section 5 and the detailed results of the experiments will be dealt with in the supplementary materials.

\section{Background}
\label{sec:headings}

\subsection{Back-propagation based CNN \& RNN Training}
Back-propagation (\cite{BP}) is a widely used algorithm for the DNN training. It computes gradient of the loss function with respect to the weights of the network by using chain rule. Let assume that the DNN consists of $L$ layers and it uses a non-linear function, $f(\cdot)$. The non-linear function should be placed after the multiplication of the input feature map, $\vo_{i}$, and corresponding weight, $\mW_{i}$. Then, the input of the next layer can be obtained as
    \begin{eqnarray}
        \vp_{i+1} = \mW_{i}\;\vo_{i}, \quad \vo_{i+1} = f(\vp_{i+1}), \quad i \in \{0, 1, \dots, L-1\}
    \end{eqnarray}
After loss calculation, the error propagation stage proceeds. In this stage, it propagates the errors from the last layer to the prior layer based on chain-rule. The error, $\ve_{i+1}$ should be generated to make the gradient of the weight, $\mW_{i}$. Detail operations of the error propagation and weight gradient stage are described as follows.
    \begin{eqnarray}
        \ve_{i} = (\mW_{i}^{T}\;\ve_{i+1}) \odot f'(\vp_{i+1}), \quad i \in \{1, 2, \dots, L-1\}
    \end{eqnarray}
    \begin{eqnarray}
        \mG_{i} = \ve_{i+1}\;\vo_{i}^{T}, \quad i \in \{0, 1, \dots, L-1\}
    \end{eqnarray}
    \begin{eqnarray}
        \mW_{i}^{new} = \mW_{i} - \eta\;\mG_{i}
    \end{eqnarray}
Equation (4) can be varied based on the optimizer such as momentum (\cite{Momentum}) or adam (\cite{Adam}).
In the CNN training, the matrix multiplication operations are substituted with the convolution operations. In addition, 180\textdegree\space flipped kernel is used instead of the transposed weight in the equation (2).

The RNN also adopts BP based training but should consider recurrent computational characteristics. During the RNN training, the RNN is unfolded through time and calculates weight gradients based on BP. Since the unfolded RNN cells have a shared weight, the weight gradient generated in every cell should be integrated to update the corresponding weight. This method is called as `Back-propagation Through Time' (BPTT, \cite{BPTT}).

BP achieved state-of-the-art performance in DNN training but it currently shows the limitation of parallel computing. As mentioned before, the error propagation should proceed from the last layer to the first layer in order. The next inference can be started after the weight update of the first layer is finished. However, it should wait until the error propagation is finished and the error reaches the first layer. This problem is generally known as the `backward locking' problem. Nowadays, many researchers tried to multi-GPU for the high-speed DNN training but the serial computational characteristic appeared in the BP becomes the main obstacle of that purpose. 

Specifically, the problem becomes critical in RNN training. Although BPTT shows great training performance compared with other optimization algorithms such as evolution strategy (\cite{ES}), it suffers from a backward locking problem and causes slow training. When the RNN has a large time step, the depth of the unfolded RNN becomes huge. Since the error propagation of the different time steps proceeds with serial-manner, the parallel processing is impossible with the BPTT and the speed of the error propagation can be limited.

\begin{figure}[t]
    \includegraphics[width=\textwidth]{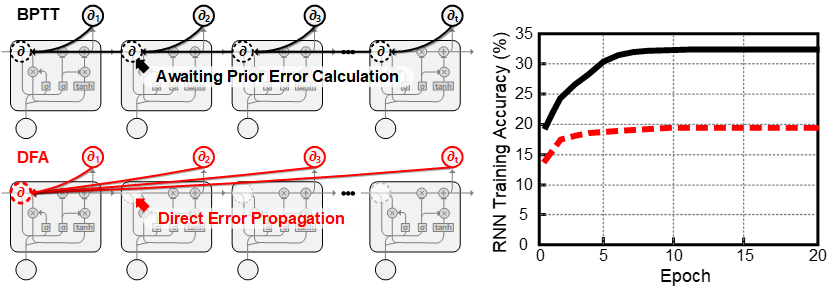}
    \caption{Back-propagation Through Time vs DFA based Error Propagation}
    \label{fig_BPTT_vs_DFA}
\end{figure}

\subsection{Direct Feedback Alignment based CNN \& RNN Training}
BP based training methodology has two main disadvantages. One of them is the weight transport problem which needs the transpose of the weights during the error propagation. The transpose reading of the weights is not appropriate for biologically plausible training and it also requires additional memory access cost. Previously, the neuron connections during the error propagation depend on forward neuron connections but the DFA got rid of this dependency by propagating errors of the last layer directly to the all intermediate layers. The backward weight, $\mD_{i}$, consists of random values but the value should be maintained even the forward weight is newly updated. With the error of the last layer, $\ve_{L}$ and related backward weight, $\mD_{i}$, the error of the intermediate layer can be calculated as 
    \begin{eqnarray}
        \ve_{i} = (\mD_{i}\;\ve_{L}) \odot f'(\vp_{i+1})
    \end{eqnarray}
The DFA is one of the biologically plausible training methods and it has the potential strength to overcome the backward locking which is the main weakness of the BP based training scenario. However, the DFA shows comparable accuracy only in the training of fully-connected layers. \cite{BDFA} tried to enlarge the DFA to CNN, but it shows high accuracy by adopting DFA only to the last few fully-connected layers. Since the portion of the fully-connected layer becomes smaller and smaller in the current CNN architecture, it is not that useful in the CNNs such as ResNet (\cite{ResNet}). 

In the conventional DFA, the connections of the backward path are considered as the fully-connected layer, so the computation methodology of the error propagation is fixed as matrix multiplication. For example, the CNN adopts convolution operations and RNN adopts recurrent computation flow in the forward path. When we apply the DFA as the error propagation method, it creates new connections between the output neurons and the intermediate neurons. And then, it allocates arbitrary synaptic weights to the newly created backward connections. Therefore, the backward path of CNN and RNN is revised as a fully-connected layer and it performs matrix multiplication instead of convolution or recurrent computation. These different computational characteristics induce low training performance and low hardware efficiency compared with BP. In CNN, the intermediate feature map placed in the front layer generally has a large size. It induces that the DFA has more computation amounts compared with the BP algorithm. When the DFA is applied to the RNN, the problem can be significantly increased due to the time step. The number of outputs can be increased if the size of the time step is large. It induces large memory and computations due to the connections related to the output neurons of the every time step. 

In summary, conventional DFA was an important step forward to biologically plausible DNN training. However, it cannot be the substitute for using BP because of the low training performance and efficiency.

\section{Our Approach} 
In this section, we will introduce the new DFA algorithm which has high scalability to various types of network architecture while maintaining the advantages of the conventional DFA algorithm. At first, we divide the DNN into small modules and apply the DFA inside the module. Second, dilated convolution and shufflenet (\cite{ShuffleNet}) based channel division will be combined with the DFA algorithm in the CNN training. Third, the upper triangular and highly sparse backward weight is adopted to improve both the accuracy and efficiency of the RNN training. At last, we propose the hybrid DFA (HDFA) which blends BP and DFA to compensate for the accuracy loss caused by DFA. 

\begin{figure}[b]
    \includegraphics[width=\textwidth]{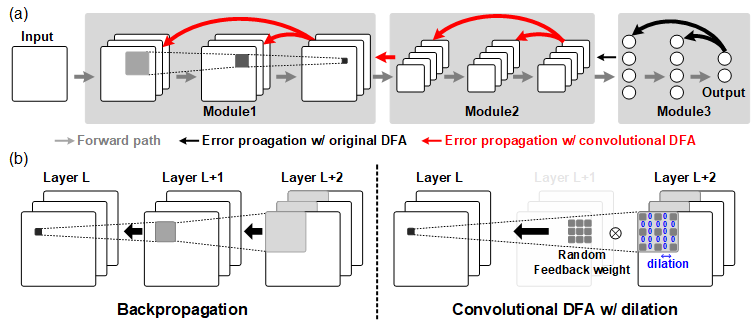}
    \caption{(a) Error Propagation Path by using DFA with Network Modularization (b) Training the Convolutional Layer by using Back-propagation (Left) and Convolutional DFA (Right)}
    \label{figure_cnn_explain}
\end{figure}
\subsection{Network Modularization}
As depicted in figure 1, the DFA based training shows much lower training accuracy compared with the BP algorithm. The main reason for the accuracy drop is a large number of layers. The DFA shows lower accuracy when it is applied to deep CNNs. \cite{BDFA} shows that using DFA only on the last few layers can achieve high training accuracy while maintaining BP based error propagation on the prior layers. Therefore, we tried to revise the original DFA algorithm by reducing the distance of the direct error propagation path.

Our proposed DFA adopts modularization which divides the network into several modules. The distance of direct connection becomes shorter than before because the direct error propagation is done only inside the module. The number of modules is fixed in RNN training but it is a design choice in the CNN training.

In the CNN training, the number of modules has a trade-off between training accuracy and a chance of parallel processing. When the number of modules is increased, the training accuracy of the DFA can be increased because the distance of the error propagation becomes shorter and the computation characteristic becomes similar to BP. In contrast, it can disturb the multi-GPU acceleration because the number of direct paths is decreased due to modularization. Let $M$ is the total number of error propagation module, then the computation of DFA is revised as follow.
    \begin{equation}
        \begin{aligned}
            \begin{cases}
            \ve_{i,j} \quad \quad \:\,   = (\mD_{i,j}\;\ve_{i,L_{i}}) \odot f'(\vp_{i,j+1}) & \text{$j\ne-1$} \\
            \ve_{i-1,L_{i-1}} = (\mD_{i,j}\;\ve_{i,L_{i}}) \odot f'(\vp_{i,0}) & \text{$j=-1$}
            \end{cases}
        \quad i \in \{1, 2, \dots, M\}, \quad j \in \{-1, 0, 1, \dots, L_{i}-1\}
        \end{aligned}
    \end{equation}
The number of layers is small in the RNN cells so we divide the RNN cell only into two modules. The first module includes both the embedding layer and the basic RNN cell. The other module contains the last fully-connected layer which decides the final output based on the result of RNN cells. The original DFA based RNN training suffers from a large computation amount because there are lots of output neurons in the last layer. Thanks to modularization, direct propagation is only performed from the RNN cell which has a much smaller number of output neurons. Moreover, this division induces accuracy improvement due to the short error propagation path.

\subsection{Computational Optimization of CNN}
\subsubsection{Dilated Convolution based Error Propagation}

Training CNN with the conventional DFA requires error propagation in a fully-connected fashion. However, as seen in figure 2 (a), a single output element is affected only by the elements inside the specific area, receptive field. Conventional DFA does not consider the receptive field but connects massive unrelated connections during the error propagation. As a result, the complexity of error propagation grows exponentially and makes it hard to train the network with the conventional DFA. In our proposed DFA, these unnecessary connections are removed by adopting the convolutional direct error propagation. As shown in figure 2 (b), error in the convolution layer is propagated by the series of convolutions during the back-propagation. In consequence, the error of the $L+2^{th}$ layer requires a widened receptive field to generate a single error element at layer $L$. In order to directly propagate error from layer $L+2$ to $L$, the same receptive field should be maintained. 

\cite{DCONV} adopted dilated convolution-based network architecture to exploit a large receptive field without increasing the size of the forward weight. The additional computation required by using an expanded receptive field can be resolved by utilizing the zero-skipping DNN hardware (\cite{DT-CNN}). Therefore, we adopted the dilated convolution for direct error propagation that occurred within the module. As shown in figure 2 (b), the convolutional DFA enlarges the receptive field of the backward weight to the required size by inserting zeros inside this backward weight. In addition to this, binary backward weight is utilized to minimize the size of the required memory to store the backward weight. In this way, the error is propagated efficiently both in terms of memory and computation while maintaining its alignment with the forward propagation. 

\subsubsection{Division of Channels during Error Propagation}

  \begin{figure}[b]
    \includegraphics[width=\textwidth]{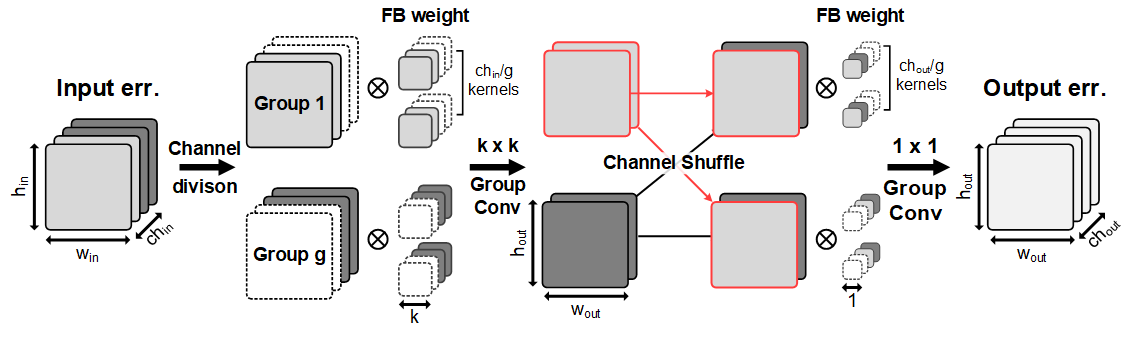}
    \caption{Group-convolution based Error Propagation Method for CNN}
    \label{figure_cnn_gconv_prop}
\end{figure}

Error propagation with dilated convolution makes the computation and memory requirement comparable to that of the back-propagation. In this work, we further optimized the error propagation computation by adopting the group convolution and channel shuffling concept used in the previous mobile-optimized network, ShuffleNet (\cite{SF})

During the group-convolution based error propagation, the errors are divided into several groups of channels, and each group is computed independently. 
Denote the error as $\ve_{i} = \begin{bmatrix} \ve^{1}_{i}, \, \ve^{2}_{i}, \, \cdots \, ,\ve^{G}_{i} \end{bmatrix}$, where $\, i \in \, \{1, 2, \cdots , L-1 \}$ represents the layer number. $\ve^{j}_{i}$ denotes error at $j^{th}$ channel group, where $\, j \in \, \{1, 2, \cdots , G \}$ represents the channel group number. 
With a $k \times k$ backward weight $\bm{D_1}$ and $1 \times 1$ backward weight $\bm{D_2}$, calculation of the error $\ve_{i}$ is formulated as below. 
    \begin{equation}
        \begin{aligned}
            \ve_{i} = (\bm{D_1} \otimes^{g} \textit{shuffle}(\bm{D_2} \otimes^{g} \ve_{L} )) \odot f'(\vp_{i+1}) 
        \end{aligned}
    \end{equation}
Group convolution between weight, $\bm{D}$, and error, $\ve$, can be expressed as follows.
    \begin{equation}
        \begin{aligned}
            \bm{D} \otimes^{g} 
            \ve = 
            \begin{bmatrix} \bm{D}^1 \otimes \ve^1, \bm{D}^2 \otimes \ve^2, \dots, \bm{D}^G \otimes \ve^G \end{bmatrix}
        \end{aligned}
    \end{equation}
Compared with the standard convolution $\bm{D} \otimes \ve$, $\bm{D}^i \otimes \ve^i$ requires $1/G^2$ times less computation by removing channel connections among groups. Therefore, group convolution involves $1/G$ less number of computation. Likewise, the size of the backward weight, $\bm{D}$, is $1/G$ times smaller thanks to the channel division.
    
As a result of modified error propagation methodology, both memory transaction and the number of computation are reduced as follows. Denote \{width, height, and number of channels\} for $\ve_{L}$ and $\ve_{i}$ as \{$w_{L}$, $h_{L}$, $ch_{L}$\} and \{$w_{i}$, $h_{i}$, $ch_+{i}$\} respectively. 
Convolutional DFA used in section 3.2.1 requires $k^2 \times ch_{L} \times w_{i} \times h_{i} \times ch_{i}$ multiplications. On the other hand, proposed group convolution based error propagation requires $k^2 \times ch_{L} \times w_{i} \times h_{i} \times ch_{i} \times (1/k^2 + ch_{L}/(ch_{i} \times G))$ multiplications. For instance, assume $1/k^2 \ll 1/G$ and $ch_{i} = ch_{L}$.
Then the low complexity error propagation method approximately reduces the memory and computation requirement by $G$ times.

\subsection{Computational Optimization of RNN}

\begin{figure}[b]
    \includegraphics[width=\textwidth]{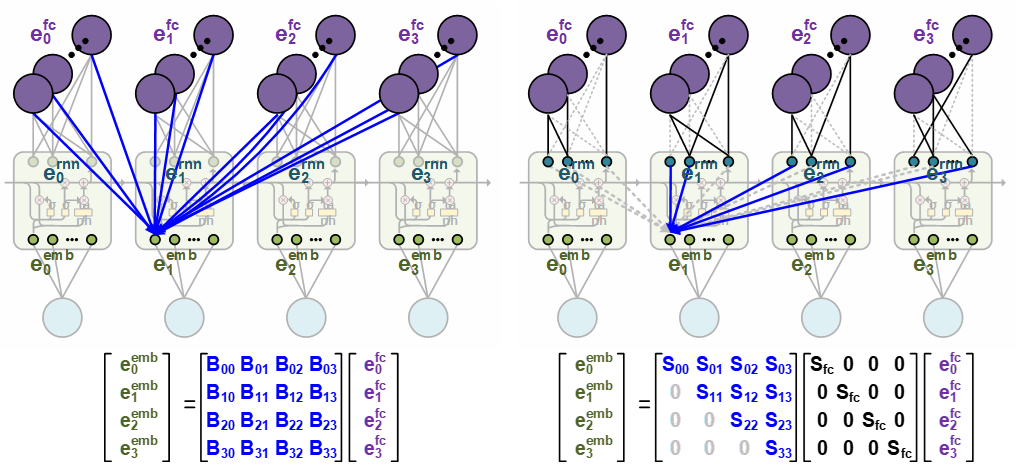}
    \caption{Proposed RNN Training Method by using DFA (Left: Original DFA, Right: Modularization + Upper Triangular Backward Weight + Sparse Backward Weight)}
    \label{figure_rnn_explain}
\end{figure}

\subsubsection{Revision of Backward Weight into Upper Triangular Matrix}
Refer to \cite{FA}, the angle between backward weight, $\bm{D}$, and the transpose of forward weight, $\mW^{T}$ becomes similar when the training proceeds. In other words, the network automatically adapts its random backward weights by modifying the corresponding forward weights. Therefore, the training accuracy can be dropped when the forward weight cannot easily predict the backward weight during the training process. 

In the conventional DFA based RNN training, all intermediate neurons are connected with the output neurons of the last layer during the error propagation. However, this learning process makes the training process more difficult because of the unnecessary backward connections. Let the $\mX_{i}$ as the input value of the RNN cell at the time step $i$. Figure 4 shows the unfolded form of the RNN computation. Final output at the time step $\vi$ is affected by the inputs, $\mX_{0}$ to $\mX_{i-1}$ which are the values of the previous and current inputs. Since the backward connection is constructed even with the future inputs ($\mX_{i+1}$, $\mX_{i+2}$, \dots) in the conventional DFA, it finally becomes the main reason for the low accuracy. 

To improve the performance of the DFA, we disconnect the backward connections which have no impact on the final output at the forward computation. The backward weight used in the DFA is modified to reflect the data dependency, and the backward weight shows the shape of the upper triangular matrix as depicted in Figure 4. Another observation is that the input values of RNN cells experience the same operations even in the next time step. The backward weight is modified to reflect the repeated operation by using the power of partial backward weight. In summary, the final backward weight is determined as follows when the number of output neurons of the RNN cell is $H$ and the total time step is $T$.
    \begin{equation}
        \begin{aligned}
        \mB_{HT\times HT}^{rnn} &=
            \begin{cases}
            \mB_{i, j}^{rnn} = power(\mD_{i,j}^{rnn}, T-j) & \text{$i\leq j$} \\
            \mB_{i, j}^{rnn} = \mO_{H\times H} & \text{$i>j$}
            \end{cases}
        \quad i, j \in \{0, 1, \dots, T-1\}
        \end{aligned}
    \end{equation}

\subsubsection{Highly-sparse and Binary Backward Weight}
The backward weight used in the RNN training becomes huge because the backward connections are constructed for every output neuron of unfolded RNN cells. These connections induce the large computation amount and storage to store the backward weight.
One of the possible solutions is the binarization introduced by \cite{BDFA}. It modifies the backward weight which has only the sign values, +1/-1. This modification shows no accuracy degradation but reduces the total memory storage by 96.9\%. It successfully reduces the memory size but does not affect the total number of computations.

The other solution is using the sparse backward weight suggested by \cite{SDFA}. \cite{SDFA} verified that the performance of the DFA is not that degraded with the backward weight including lots of zeros. The sparse weight can reduce the overall necessary computations during the error propagation. Zero-skipping DNN accelerator architecture such as EIE (\cite{EIE}) can be efficiently used to improve the processing speed of the error propagation stage. We finally combine both the binarized and highly-sparse backward weight to optimize both memory storage and computations during the RNN training. The effect of the binarization and sparse backward weight will be discussed in section 4. When the sparse bitmap mask is denoted as $R$, the final backward weight is calculated as follows.
    \begin{eqnarray}
        \mS^{fc} = sign(\mD^{fc} \odot R^{fc}), \quad \ve_{i}^{rnn} = (\mS^{fc}\;\ve_{i}^{fc}) \odot f'(\vp^{fc})
    \end{eqnarray}
    \begin{equation}
        \begin{aligned}
        \mS_{HT\times HT}^{rnn} &=
            \begin{cases}
            \mS_{i, j}^{rnn} = sign(power(\mD_{i,j}^{rnn}, T-j) \odot R_{j}^{rnn}) & \text{$i\leq j$} \\
            \mS_{i, j}^{rnn} = \mO_{H\times H} & \text{$i>j$}
            \end{cases}
        , \quad i, j \in \{0, 1, \dots, T-1\}
        \end{aligned}
    \end{equation}
    \begin{eqnarray}
        \ve^{emb} = \begin{bmatrix} 
            \ve_{0}^{emb} \\
            \ve_{1}^{emb} \\
            \vdots \\
            \ve_{T-1}^{emb}
        \end{bmatrix}, \quad
        \ve^{rnn} = \begin{bmatrix} 
            \ve_{0}^{rnn} \\
            \ve_{1}^{rnn} \\
            \vdots \\
            \ve_{T-1}^{rnn}
        \end{bmatrix}, \quad
        \quad \ve^{emb} = (\mS^{rnn}\;\ve^{rnn}) \odot f'(\vp^{rnn})
    \end{eqnarray}

\subsection{Hybrid Direct Feedback Alignment}
    
    \begin{algorithm}[b]
    \caption{Custom Optimizer of the Hybrid Direct Feedback Alignment}
    \label{hydfa}
    \begin{algorithmic}
            \Require{Learning rate $\eta$, momentum parameter $\alpha$, mix parameter $\gamma$, BP ratio $p$}
            \Require{Initial parameter $\bm{\theta}$, Initial BP velocity $\bm{v}^{BP}$, Initial DFA velocity $\bm{v}^{DFA}$}
            \Require{Objective function $\bm{f}(\bm{x};\bm{\theta})$ given input $\bm{x}$ and parameters $\bm{\theta}$}
            \\ Initialize network parameters $\bm{\theta}$
            \\ Initialize momentum for BP $\bm{v}^{BP} \leftarrow 0$ and momentum for DFA $\bm{v}^{DFA} \leftarrow 0$

            \While {$\bm{\theta}$ not converged}
            \State Sample a minibatch of $m$ examples from the training set $\{x^{(1)},\cdots,x^{(m)}\}$ with corresponding targets $y^{(i)}$ 
            \State Set $rand = random(0, \, 1)$
                \If {$rand \leq p$}
                \State
                    $\bm{g} \leftarrow \frac{1}{m} \nabla_{\bm{\theta}}^{BP} \sum_{i} L(f(\bm{x}^{(i)}, \bm{\theta}), \bm{y}^{(i)})$
                    \Comment{Compute gradient through BP}
                \State
                    $\bm{v}^{BP} \leftarrow \bm{v}^{BP} + \alpha \bm{g}$
                    \Comment{Compute momentum update}
                \State
                    $\bm{\theta} \leftarrow \bm{\theta} + \eta \bm{v}^{BP}$
                    \Comment{Update parameter}
                \Else
                \State
                    $\bm{g} \leftarrow \frac{1}{m} \nabla_{\bm{\theta}}^{DFA} \sum_{i} L(f(\bm{x}^{(i)}, \bm{\theta}), \bm{y}^{(i)})$
                    \Comment{Compute gradient through DFA in Section 3.2.2}
                \State
                    $\bm{v}^{DFA} \leftarrow \bm{v}^{DFA} + \alpha \bm{g}$
                    \Comment{Compute momentum update}
                \State
                    $\bm{u} \leftarrow \gamma \bm{v}^{DFA} + (1-\gamma) \bm{v}^{BP}$
                    \Comment{Mix DFA momentum with BP momentum}
                \State
                    $\bm{\theta} \leftarrow \bm{\theta} + \eta \bm{u}$
                    \Comment{Update parameter}
                \EndIf
            \EndWhile
            \\ \Return {$\bm{\theta}$}
    \end{algorithmic}
    \end{algorithm}
    
    We can get higher accuracy by modifying the error propagation methods as explained in section 3.1 to 3.3. However, the accuracy of the DFA in both CNN and RNN is still below the one trained with the BP algorithm. Hybrid DFA (HDFA) is newly proposed to be comparable with the BP algorithm in terms of the training accuracy.
    
    In the case of RNN training, the HDFA works as follows. The error propagation method is randomly determined to be BP or DFA at each iteration. The weight gradients generated by two different error propagation methods are integrated into the momentum term of adam optimizer (\cite{Adam}). We found that the gradient occasionally generated by BP accelerates the convergence of the DFA algorithm and it finally helps the DFA algorithm to have higher accuracy. 
    
    The optimizer used in CNN training is different from conventional optimizers such as momentum or adam. The HDFA based CNN training with the conventional optimizers fails to accomplish BP-level accuracy. To train the CNN with the proposed HDFA, original momentum optimizer (\cite{Momentum}) should be modified as described in algorithm 1. Unlike the conventional optimizers, our proposed optimizer distinguishes the momentum terms of the DFA from the BP. We denote the momentum of the BP as $\bm{v}^{BP}$, and the DFA as $\bm{v}^{DFA}$. The error generated by each algorithm affects only its own momentum value. The frequency of the BP algorithm is determined by the hyper-parameter, BP ratio ($p$). When the error is propagated through the DFA, both $\bm{v}^{DFA}$ and $\bm{v}^{BP}$ is considered to make the final weight gradient. Thanks to the revision of optimizer, the HDFA can achieve high training accuracy even in the CNN training scenario.

\section{Experiments}
In this section, we compared the training accuracy of the conventional BP, DFA, and the proposed new DFA algorithm, HDFA. In the CNN training scenario, we used the CIFAR-10 dataset (\cite{CIFAR}) with the random cropping and random horizontal flip data augmentation method. We verified the performance in the various CNNs such as VGG16 (\cite{VGG}) and ResNet (\cite{ResNet}). The simulation was based on mini-batch gradient descent with the batch size, 128. The gradient optimizer is based on momentum (\cite{Momentum}), but it is revised based on algorithm 1. 

In the RNN training, the WikiText-2 dataset (\cite{wikitext}) is used to predict the English word sequence. We tested three different types of RNN cells: 1) first of all, vanilla RNN which only contains the tanh activation function, 2) long short-term memory (LSTM, \cite{LSTM}), and 3) gated recurrent unit (GRU, \cite{GRU}). All RNN cells receive 200 embedding values and extract 200 values of hidden state while it adopts a 35-time step and 20 batch size. The optimizer of the RNN is decided as adam (\cite{Adam}) because it shows a higher accuracy compared with the momentum.

\subsection{CNN Training Result}

\begin{table}[b]
\caption{CNN Training and Test Accuracy [\%]}
\label{table1}
\setlength\extrarowheight{1pt}
\begin{center}
\begin{tabular}{cccccccc}
\hline
 &
   &
  \multicolumn{6}{c}{\textbf{Network}} \\ \cline{3-8} 
 &
  \multirow{2}{*}{\textbf{\begin{tabular}[c]{@{}c@{}}Channel\\ Groups\end{tabular}}} &
  \multicolumn{2}{c}{\textbf{VGG16}} &
  \multicolumn{2}{c}{\textbf{R-18}} &
  \multicolumn{2}{c}{\textbf{R-50}} \\ \cline{3-8} 
 &
   &
  \textbf{Train} &
  \textbf{Test} &
  \textbf{Train} &
  \textbf{Test} &
  \textbf{Train} &
  \textbf{Test} \\ \hline
\textbf{BP} &
  1 &
  \textbf{99.91} &
  \textbf{92.68} &
  \textbf{99.94} &
  \textbf{92.74} &
  \textbf{99.97} &
  \textbf{93.07} \\ \hline
\textbf{Original DFA} &
  - &
  86.66 &
  77.78 &
  66.53 &
  66.19 &
  68.95 &
  71.92 \\ \hline
\textbf{Convolutional DFA} &
  1 &
  84.84 &
  78.18 &
  81.79 &
  76.55 &
  - &
  - \\ \hline
\textbf{+Modularization} &
  1 &
  84.96 &
  78.56 &
  88.99 &
  79.60 &
  82.81 &
  77.99 \\ \hline
\textbf{+Dilation} &
  1 &
  81.22 &
  77.41 &
  88.05 &
  78.09 &
  82.44 &
  77.27 \\ \hline
\multirow{2}{*}{\textbf{\begin{tabular}[c]{@{}c@{}}+Group Convolution\\ \&Channel Shuffling\end{tabular}}} &
  4 &
  82.60 &
  76.85 &
  88.99 &
  79.09 &
  86.57 &
  80.4 \\
 &
  8 &
  80.75 &
  77.5 &
  87.94 &
  79.03 &
  87.17 &
  81.36 \\ \hline
\multirow{2}{*}{\textbf{\begin{tabular}[c]{@{}c@{}}+BP/BDFA\\ =0.1/0.9\end{tabular}}} &
  4 &
  93.39 &
  87.92 &
  94.05 &
  88.35 &
  94.37 &
  88.55 \\
 &
  8 &
  93.64 &
  87.98 &
  93.39 &
  88.29 &
  93.71 &
  87.93 \\ \hline
\multirow{2}{*}{\textbf{\begin{tabular}[c]{@{}c@{}}+BP/BDFA\\ =0.3/0.7\end{tabular}}} &
  4 &
  97.45 &
  90.15 &
  97.66 &
  91.10 &
  97.41 &
  90.13 \\
 &
  8 &
  97.45 &
  90.28 &
  97.60 &
  90.83 &
  98.05 &
  91.51 \\ \hline
\multirow{2}{*}{\textbf{\begin{tabular}[c]{@{}c@{}}+BP/BDFA\\ =0.5/0.5\end{tabular}}} &
  4 &
  98.90 &
  91.10 &
  99.18 &
  91.78 &
  98.87 &
  91.42
   \\
 &
  8 &
  99.00 &
  91.46 &
  99.27 &
  92.18 &
  99.57 &
  92.21 \\ \hline
\end{tabular}
\end{center}
\end{table}

The training performance of the conventional and proposed algorithm is summarized in Table 1. In conventional DFA based training, it shows poor accuracy when the number of layers is increased. The network modularization helps the DFA to have higher accuracy since it reduces the distance of the direct error propagation path. The modification of the error propagation methodology affects training performance in different ways. In the VGG-16, the training accuracy is getting lower when the error propagation method becomes simpler due to the dilated and group convolutions. On the other hand, the modifications make the ResNets have higher accuracy. In this experiment, we verified that the residual connection of the ResNet helps the robustness to the DFA based training and shows the best result when it is combined with the simpler error calculations. Although VGG-16 shows lower accuracy compared with the original DFA, the accuracy degradation can be compensated with the hybrid training algorithms. We examined three different BP/DFA ratios by adopting HDFA introduced in Section 3.4. The HDFA finally raises accuracy as similar to the final result of BP. 

Thanks to the modification of error calculation and HDFA, we can get the BP-level high accuracy with less hardware cost. Table 2 shows the size of the memory transactions and the number of operations required during the error propagation stage. Since the number of the class is small in the CIFAR-10 dataset, both conventional and proposed DFA algorithm, shows a smaller number of operations during the error propagation stage. (The detailed relationship between class number and operation will be discussed in Appendix A.1) Instead, conventional DFA requires large backward weight memory transactions. That's because BP shares weights for multiple channels but DFA allocates different weights for all connections. Dilated and group-convolution based error propagation solve this problem and finally, the proposed DFA algorithm shows 99.0\% lower memory transaction of backward weight data and 60.1\% smaller number of operations compared with the BP. 

\begin{table}[t]
    \caption{Comparison of Hardware Efficiency in the CNN Training}
    \begin{subtable}{.5\linewidth}
      \centering
        \caption{Memory Transaction during EP [MB]}
        \begin{tabular}{ccccc}
        \hline
         &
          \multirow{2}{*}{\textbf{\begin{tabular}[c]{@{}c@{}}Channel\\ Groups\end{tabular}}} &
          \multicolumn{3}{c}{\textbf{Network}} \\ \cline{3-5} 
                                              &   & \textbf{VGG16} & \textbf{R-18} & \textbf{R-50} \\ \hline
        \textbf{BP}                           & 1 & \textbf{56.13} & \textbf{42.64}    & \textbf{53.5}     \\ \hline
        \textbf{DFA}                          & - & 6.97           & 25.39             & 149.83            \\ \hline
        \multirow{2}{*}{\textbf{Revised DFA}} & 4 & 0.54           & 0.43              & 1.06              \\
                                              & 8 & 0.27           & 0.23              & 0.54              \\ \hline
        \multirow{2}{*}{\textbf{\begin{tabular}[c]{@{}c@{}}+BP/BDFA\\ =0.1/0.9\end{tabular}}} &
          4 &
          6.10 &
          4.65 &
          6.30 \\
                                              & 8 & 5.87           & 4.47              & 5.84              \\ \hline
        \multirow{2}{*}{\textbf{\begin{tabular}[c]{@{}c@{}}+BP/BDFA\\ =0.3/0.7\end{tabular}}} &
          4 &
          17.22 &
          13.09 &
          16.79 \\
                                              & 8 & 17.03          & 12.95             & 16.43             \\ \hline
        \end{tabular}
    \end{subtable}%
    \begin{subtable}{.5\linewidth}
      \centering
        \caption{The Number of Operations during EP [GOP]}
        \begin{tabular}{ccccc}
        \hline
         &
          \multirow{2}{*}{\textbf{\begin{tabular}[c]{@{}c@{}}Channel\\ Groups\end{tabular}}} &
          \multicolumn{3}{c}{\textbf{Network}} \\ \cline{3-5} 
                                              &   & \textbf{VGG16} & \textbf{R-18} & \textbf{R-50} \\ \hline
        \textbf{BP}                           & 1 & \textbf{0.62}  & \textbf{1.48}     & \textbf{1.53}     \\ \hline
        \textbf{DFA}                          & - & 0.003          & 0.01              & 0.08              \\ \hline
        \multirow{2}{*}{\textbf{Revised DFA}} & 4 & 0.26           & 0.56              & 1.17              \\
                                              & 8 & 0.13           & 0.31              & 0.61              \\ \hline
        \multirow{2}{*}{\textbf{\begin{tabular}[c]{@{}c@{}}+BP/BDFA\\ =0.1/0.9\end{tabular}}} &
          4 &
          0.3 &
          0.65 &
          1.2 \\
                                              & 8 & 0.18           & 0.42              & 0.7               \\ \hline
        \multirow{2}{*}{\textbf{\begin{tabular}[c]{@{}c@{}}+BP/BDFA\\ =0.3/0.7\end{tabular}}} &
          4 &
          0.37 &
          0.84 &
          1.28 \\
                                              & 8 & 0.28           & 0.66              & 0.88              \\ \hline
        \end{tabular}
    \end{subtable} 
\end{table}

\subsection{RNN Training Result}

\begin{table}[hbt!]
\caption{RNN Training and Test Accuracy [\%]}
\label{table2}
\setlength\extrarowheight{1pt}
\begin{center}
\begin{tabular}{|C|C|C|C|C|C|C|C|}
    \hline
    \multicolumn{1}{c}{\multirow{3}{*}{}} &
    \multicolumn{1}{c}{\multirow{3}{*}{}} &
    \multicolumn{6}{c}{\bf Network} \\
    \cline{3-8}
    \multicolumn{1}{c}{} &
    \multicolumn{1}{c}{\bf BW} &
    \multicolumn{2}{c}{\bf RNN} &
    \multicolumn{2}{c}{\bf LSTM} &
    \multicolumn{2}{c}{\bf GRU} \\
    \cline{3-8}
    \multicolumn{1}{c}{} &
    \multicolumn{1}{c}{\bf Sparsity} &
    \multicolumn{1}{c}{\bf Train} &
    \multicolumn{1}{c}{\bf Test} &
    \multicolumn{1}{c}{\bf Train} &
    \multicolumn{1}{c}{\bf Test} &
    \multicolumn{1}{c}{\bf Train} &
    \multicolumn{1}{c}{\bf Test} \\
    \hline
    \multicolumn{1}{c}{\bf BP} &
    \multicolumn{1}{c}{-} &
    \multicolumn{1}{c}{\bf 29.55} &
    \multicolumn{1}{c}{\bf 23.00} &
    \multicolumn{1}{c}{\bf 32.24} &
    \multicolumn{1}{c}{\bf 24.48} &
    \multicolumn{1}{c}{\bf 32.53} &
    \multicolumn{1}{c}{\bf 24.52} \\
    \hline
    \multicolumn{1}{c}{\bf Original DFA} &
    \multicolumn{1}{c}{-} &
    \multicolumn{1}{c}{20.69} &
    \multicolumn{1}{c}{18.35} &
    \multicolumn{1}{c}{20.84} &
    \multicolumn{1}{c}{19.19} &
    \multicolumn{1}{c}{19.62} &
    \multicolumn{1}{c}{18.15} \\
    \hline
    \multicolumn{1}{c}{\bf + Triangular BW} &
    \multicolumn{1}{c}{-} &
    \multicolumn{1}{c}{22.93} &
    \multicolumn{1}{c}{20.33} &
    \multicolumn{1}{c}{23.30} &
    \multicolumn{1}{c}{20.70} &
    \multicolumn{1}{c}{23.04} &
    \multicolumn{1}{c}{20.65} \\
    \hline
    \multicolumn{1}{c}{\bf + BW Init} &
    \multicolumn{1}{c}{-} &
    \multicolumn{1}{c}{24.30} &
    \multicolumn{1}{c}{21.00} &
    \multicolumn{1}{c}{24.54} &
    \multicolumn{1}{c}{21.19} &
    \multicolumn{1}{c}{24.77} &
    \multicolumn{1}{c}{21.35} \\
    \hline
    \multicolumn{1}{c}{\multirow{4}{*}{}} &
    \multicolumn{1}{c}{0\%} &
    \multicolumn{1}{c}{27.69} &
    \multicolumn{1}{c}{22.48} &
    \multicolumn{1}{c}{29.50} &
    \multicolumn{1}{c}{23.62} &
    \multicolumn{1}{c}{30.54} &
    \multicolumn{1}{c}{23.70} \\
    \multicolumn{1}{c}{\bf +BP/DFA*} &
    \multicolumn{1}{c}{80\%} &
    \multicolumn{1}{c}{27.08} &
    \multicolumn{1}{c}{22.66} &
    \multicolumn{1}{c}{29.28} &
    \multicolumn{1}{c}{24.03} &
    \multicolumn{1}{c}{30.04} &
    \multicolumn{1}{c}{24.16} \\
    \multicolumn{1}{c}{\bf = 0.1/0.9} &
    \multicolumn{1}{c}{95\%} &
    \multicolumn{1}{c}{26.85} &
    \multicolumn{1}{c}{22.47} &
    \multicolumn{1}{c}{28.77} &
    \multicolumn{1}{c}{23.99} &
    \multicolumn{1}{c}{29.84} &
    \multicolumn{1}{c}{24.18} \\
    \multicolumn{1}{c}{} &
    \multicolumn{1}{c}{98\%} &
    \multicolumn{1}{c}{26.16} &
    \multicolumn{1}{c}{22.41} &
    \multicolumn{1}{c}{28.61} &
    \multicolumn{1}{c}{24.04} &
    \multicolumn{1}{c}{29.88} &
    \multicolumn{1}{c}{24.25} \\
    \hline
    \multicolumn{1}{c}{\multirow{4}{*}{}} &
    \multicolumn{1}{c}{0\%} &
    \multicolumn{1}{c}{28.30} &
    \multicolumn{1}{c}{22.77} &
    \multicolumn{1}{c}{30.65} &
    \multicolumn{1}{c}{24.11} &
    \multicolumn{1}{c}{31.28} &
    \multicolumn{1}{c}{24.18} \\
    \multicolumn{1}{c}{\bf +BP/DFA*} &
    \multicolumn{1}{c}{80\%} &
    \multicolumn{1}{c}{28.02} &
    \multicolumn{1}{c}{22.79} &
    \multicolumn{1}{c}{30.57} &
    \multicolumn{1}{c}{24.33} &
    \multicolumn{1}{c}{30.95} &
    \multicolumn{1}{c}{24.39} \\
    \multicolumn{1}{c}{\bf = 0.2/0.8} &
    \multicolumn{1}{c}{95\%} &
    \multicolumn{1}{c}{27.27} &
    \multicolumn{1}{c}{22.73} &
    \multicolumn{1}{c}{30.23} &
    \multicolumn{1}{c}{24.36} &
    \multicolumn{1}{c}{31.10} &
    \multicolumn{1}{c}{24.41} \\
    \multicolumn{1}{c}{} &
    \multicolumn{1}{c}{98\%} &
    \multicolumn{1}{c}{26.50} &
    \multicolumn{1}{c}{22.71} &
    \multicolumn{1}{c}{30.42} &
    \multicolumn{1}{c}{24.32} &
    \multicolumn{1}{c}{31.02} &
    \multicolumn{1}{c}{24.46} \\
    \hline
    \multicolumn{1}{c}{\multirow{4}{*}{}} &
    \multicolumn{1}{c}{0\%} &
    \multicolumn{1}{c}{27.75} &
    \multicolumn{1}{c}{22.55} &
    \multicolumn{1}{c}{29.70} &
    \multicolumn{1}{c}{23.70} &
    \multicolumn{1}{c}{30.02} &
    \multicolumn{1}{c}{23.80} \\
    \multicolumn{1}{c}{\bf +BP/BDFA*} &
    \multicolumn{1}{c}{80\%} &
    \multicolumn{1}{c}{27.02} &
    \multicolumn{1}{c}{22.59} &
    \multicolumn{1}{c}{29.50} &
    \multicolumn{1}{c}{23.98} &
    \multicolumn{1}{c}{30.60} &
    \multicolumn{1}{c}{24.18} \\
    \multicolumn{1}{c}{\bf = 0.1/0.9} &
    \multicolumn{1}{c}{95\%} &
    \multicolumn{1}{c}{26.84} &
    \multicolumn{1}{c}{22.48} &
    \multicolumn{1}{c}{29.30} &
    \multicolumn{1}{c}{24.11} &
    \multicolumn{1}{c}{30.28} &
    \multicolumn{1}{c}{24.23} \\
    \multicolumn{1}{c}{} &
    \multicolumn{1}{c}{98\%} &
    \multicolumn{1}{c}{26.12} &
    \multicolumn{1}{c}{22.42} &
    \multicolumn{1}{c}{28.10} &
    \multicolumn{1}{c}{23.97} &
    \multicolumn{1}{c}{30.19} &
    \multicolumn{1}{c}{24.23} \\
    \hline
    \multicolumn{1}{c}{\multirow{4}{*}{}} &
    \multicolumn{1}{c}{0\%} &
    \multicolumn{1}{c}{28.08} &
    \multicolumn{1}{c}{22.83} &
    \multicolumn{1}{c}{30.77} &
    \multicolumn{1}{c}{24.18} &
    \multicolumn{1}{c}{31.10} &
    \multicolumn{1}{c}{24.22} \\
    \multicolumn{1}{c}{\bf +BP/BDFA*} &
    \multicolumn{1}{c}{80\%} &
    \multicolumn{1}{c}{27.34} &
    \multicolumn{1}{c}{22.79} &
    \multicolumn{1}{c}{30.07} &
    \multicolumn{1}{c}{24.25} &
    \multicolumn{1}{c}{30.36} &
    \multicolumn{1}{c}{24.42} \\
    \multicolumn{1}{c}{\bf = 0.2/0.8} &
    \multicolumn{1}{c}{95\%} &
    \multicolumn{1}{c}{27.66} &
    \multicolumn{1}{c}{22.76} &
    \multicolumn{1}{c}{30.26} &
    \multicolumn{1}{c}{24.27} &
    \multicolumn{1}{c}{30.97} &
    \multicolumn{1}{c}{24.46} \\
    \multicolumn{1}{c}{} &
    \multicolumn{1}{c}{98\%} &
    \multicolumn{1}{c}{26.84} &
    \multicolumn{1}{c}{22.63} &
    \multicolumn{1}{c}{29.89} &
    \multicolumn{1}{c}{24.37} &
    \multicolumn{1}{c}{31.14} &
    \multicolumn{1}{c}{24.41} \\
    \hline
\end{tabular}
\end{center}
\end{table}

Table 3 summarizes the overall training performance of the conventional and proposed algorithms. Conventional DFA shows the worst accuracy but the performance of the DFA is improved with the upper triangular backward weight and backward weight initialization suggested by \cite{BDFA}. The revision of the DFA algorithm increases the training and test accuracy by a maximum of 5.2\%p and 3.2\%p. The performance of the DFA is further improved when the DFA is blended with BP. As increasing the ratio of BP, we can compensate for the accuracy degradation and achieve similar test accuracy compared with BP. 

One of the key observations is that test accuracy appeared in the HDFA is similar to the BP even though the training accuracy still appears to be lower. This observation is the same phenomenon shown in the previous DFA training paper, \cite{BDFA} and it shows that the DFA has the potential of overfitting prevention. When the sparsity of the backward weight is increased, the training accuracy is decreased while it maintains the test accuracy.

Table 4 shows the hardware efficiency of the DFA algorithm during the RNN training. We measure the predicted memory transaction amount and the number of operations appeared during the error propagation stage. Since the amount of memory transaction can be varied depending on its encoding and decoding method, we calculate the memory transaction by counting the number of nonzero values. We assume that the binarization does not affect the total number of operations and count only the nonzero values as the required computation amount. Upper triangular backward weight increases the portion of zero values by disconnecting connections between outputs and unrelated input neurons. It reduces the overall memory transaction and the number of computations by 44.6\% compared to the conventional DFA. Thanks to the backward weight binarization, we can compress the weight significantly while maintaining its training performance. After considering the sparse backward weight, both memory transactions and the nonzero operations become smaller compared with the BP approach. When the 98\% sparsity is used for backward weight, it shows 87.9\% lower memory transaction and the number of operations compared with the BP. 

\begin{table}[t]
    \caption{Comparison of Hardware Efficiency in the RNN Training}
    \begin{subtable}{.5\linewidth}
      \centering
        \caption{Memory Transaction during EP [MB]}
        \begin{tabular}{|C{0.8cm}|C{0.8cm}|C{0.8cm}|C{0.8cm}|C{0.8cm}|}
            \hline
            \multicolumn{1}{c}{} &
            \multicolumn{1}{c}{\bf BW} &
            \multicolumn{3}{c}{\bf Network} \\
            \cline{3-5}
            \multicolumn{1}{c}{} &
            \multicolumn{1}{c}{\bf Sparsity} &
            \multicolumn{1}{c}{\bf RNN} &
            \multicolumn{1}{c}{\bf LSTM} &
            \multicolumn{1}{c}{\bf GRU} \\
            \hline
            \multicolumn{1}{c}{\bf BP} &
            \multicolumn{1}{c}{-} &
            \multicolumn{1}{c}{\bf 26.94} &
            \multicolumn{1}{c}{\bf 27.90} &
            \multicolumn{1}{c}{\bf 27.58} \\
            \hline
            \multicolumn{1}{c}{\bf DFA} &
            \multicolumn{1}{c}{-} &
            \multicolumn{1}{c}{1863.6} &
            \multicolumn{1}{c}{1863.6} &
            \multicolumn{1}{c}{1863.6} \\
            \hline
            \multicolumn{1}{c}{\bf Revised DFA} &
            \multicolumn{1}{c}{-} &
            \multicolumn{1}{c}{1032.6} &
            \multicolumn{1}{c}{1032.6} &
            \multicolumn{1}{c}{1032.6} \\
            \hline
            \multicolumn{1}{c}{\multirow{4}{*}{}} &
            \multicolumn{1}{c}{0\%} &
            \multicolumn{1}{c}{31.72} &
            \multicolumn{1}{c}{31.82} &
            \multicolumn{1}{c}{31.79} \\
            \multicolumn{1}{c}{\bf +BP/BDFA} &
            \multicolumn{1}{c}{80\%} &
            \multicolumn{1}{c}{8.49} &
            \multicolumn{1}{c}{8.59} &
            \multicolumn{1}{c}{8.56} \\
            \multicolumn{1}{c}{\bf = 0.1/0.9} &
            \multicolumn{1}{c}{95\%} &
            \multicolumn{1}{c}{4.14} &
            \multicolumn{1}{c}{4.23} &
            \multicolumn{1}{c}{4.20} \\
            \multicolumn{1}{c}{} &
            \multicolumn{1}{c}{98\%} &
            \multicolumn{1}{c}{3.27} &
            \multicolumn{1}{c}{3.36} &
            \multicolumn{1}{c}{3.33} \\
            \hline
            \multicolumn{1}{c}{\multirow{4}{*}{}} &
            \multicolumn{1}{c}{0\%} &
            \multicolumn{1}{c}{31.19} &
            \multicolumn{1}{c}{31.38} &
            \multicolumn{1}{c}{31.32} \\
            \multicolumn{1}{c}{\bf +BP/BDFA} &
            \multicolumn{1}{c}{80\%} &
            \multicolumn{1}{c}{10.54} &
            \multicolumn{1}{c}{10.74} &
            \multicolumn{1}{c}{10.67} \\
            \multicolumn{1}{c}{\bf = 0.2/0.8} &
            \multicolumn{1}{c}{95\%} &
            \multicolumn{1}{c}{6.67} &
            \multicolumn{1}{c}{6.86} &
            \multicolumn{1}{c}{6.80} \\
            \multicolumn{1}{c}{} &
            \multicolumn{1}{c}{98\%} &
            \multicolumn{1}{c}{5.90} &
            \multicolumn{1}{c}{6.09} &
            \multicolumn{1}{c}{6.02} \\
            \hline
        \end{tabular}
    \end{subtable}%
    \begin{subtable}{.5\linewidth}
      \centering
        \caption{The Number of Operations during EP [GOP]}
        \begin{tabular}{|C{0.8cm}|C{0.8cm}|C{0.8cm}|C{0.8cm}|C{0.8cm}|}
            \hline
            \multicolumn{1}{c}{} &
            \multicolumn{1}{c}{\bf BW} &
            \multicolumn{3}{c}{\bf Network} \\
            \cline{3-5}
            \multicolumn{1}{c}{} &
            \multicolumn{1}{c}{\bf Sparsity} &
            \multicolumn{1}{c}{\bf RNN} &
            \multicolumn{1}{c}{\bf LSTM} &
            \multicolumn{1}{c}{\bf GRU} \\
            \hline
            \multicolumn{1}{c}{\bf BP} &
            \multicolumn{1}{c}{-} &
            \multicolumn{1}{c}{\bf 9.4} &
            \multicolumn{1}{c}{\bf 9.7} &
            \multicolumn{1}{c}{\bf 9.6} \\
            \hline
            \multicolumn{1}{c}{\bf DFA} &
            \multicolumn{1}{c}{-} &
            \multicolumn{1}{c}{18.6} &
            \multicolumn{1}{c}{18.6} &
            \multicolumn{1}{c}{18.6} \\
            \hline
            \multicolumn{1}{c}{\bf Revised DFA} &
            \multicolumn{1}{c}{-} &
            \multicolumn{1}{c}{10.3} &
            \multicolumn{1}{c}{10.3} &
            \multicolumn{1}{c}{10.3} \\
            \hline
            \multicolumn{1}{c}{\multirow{4}{*}{}} &
            \multicolumn{1}{c}{0\%} &
            \multicolumn{1}{c}{10.2} &
            \multicolumn{1}{c}{10.2} &
            \multicolumn{1}{c}{10.2} \\
            \multicolumn{1}{c}{\bf +BP/BDFA} &
            \multicolumn{1}{c}{80\%} &
            \multicolumn{1}{c}{2.7} &
            \multicolumn{1}{c}{2.8} &
            \multicolumn{1}{c}{2.8} \\
            \multicolumn{1}{c}{\bf = 0.1/0.9} &
            \multicolumn{1}{c}{95\%} &
            \multicolumn{1}{c}{1.4} &
            \multicolumn{1}{c}{1.4} &
            \multicolumn{1}{c}{1.4} \\
            \multicolumn{1}{c}{} &
            \multicolumn{1}{c}{98\%} &
            \multicolumn{1}{c}{1.1} &
            \multicolumn{1}{c}{1.1} &
            \multicolumn{1}{c}{1.1} \\
            \hline
            \multicolumn{1}{c}{\multirow{4}{*}{}} &
            \multicolumn{1}{c}{0\%} &
            \multicolumn{1}{c}{10.1} &
            \multicolumn{1}{c}{10.1} &
            \multicolumn{1}{c}{10.1} \\
            \multicolumn{1}{c}{\bf +BP/BDFA} &
            \multicolumn{1}{c}{80\%} &
            \multicolumn{1}{c}{3.5} &
            \multicolumn{1}{c}{3.5} &
            \multicolumn{1}{c}{3.5} \\
            \multicolumn{1}{c}{\bf = 0.2/0.8} &
            \multicolumn{1}{c}{95\%} &
            \multicolumn{1}{c}{2.2} &
            \multicolumn{1}{c}{2.3} &
            \multicolumn{1}{c}{2.3} \\
            \multicolumn{1}{c}{} &
            \multicolumn{1}{c}{98\%} &
            \multicolumn{1}{c}{2.0} &
            \multicolumn{1}{c}{2.1} &
            \multicolumn{1}{c}{2.0} \\
            \hline
        \end{tabular}
    \end{subtable} 
\end{table}

\subsection{Summary of the HDFA based DNN Training}
According to the CNN and RNN training experiments, we verified that the proposed HDFA training algorithm can achieve BP-level training performance but shows the lowest hardware cost to accelerate the error propagation stage. Figure 6 shows a summary of the training performance by considering both accuracy and the hardware cost. As shown in figure 6, the Conventional DFA algorithm shows low accuracy even with the high hardware cost, so it is not considered as the alternative to BP. With the revision of the DFA algorithm, we can remove the unnecessary connections and improve the accuracy compared with the conventional algorithms, but it still has the low test accuracy. HDFA with binarized and sparse backward weight finally achieves BP-level test accuracy with the lowest memory transaction and computations. 

Proposed DFA is the trial of biologically plausible DNN training and achieves high accuracy as same as the BP algorithm. Moreover, it shows fewer computation amounts compared with the conventional error propagation methodology. Since the DFA is free from the backward locking problem, it also has an advantage of parallel processing. In the CNN training, we can compute errors of multiple layers in parallel. In a single layer, the error propagation can be computed in parallel because the group convolution breaks the data dependency that occurred by the channel. In summary, the proposed DFA enables both layer-level and channel-level parallel processing by using multi-GPU.

In the RNN training, the advantage of the DFA can be magnified. In the BPTT based training flow, the error propagation suffers from a backward locking problem because the errors of the prior time step can be calculated only after the posterior error calculation is completed. The DFA directly calculates all time-step errors at once, so the error propagation can be accelerated with the multi-GPU processing. In our experiment setup, error propagation speed can be improved 35 $\times$ faster than BP even without considering zeros that appeared in the backward weights.

\begin{figure}[t]
\includegraphics[width=\textwidth]{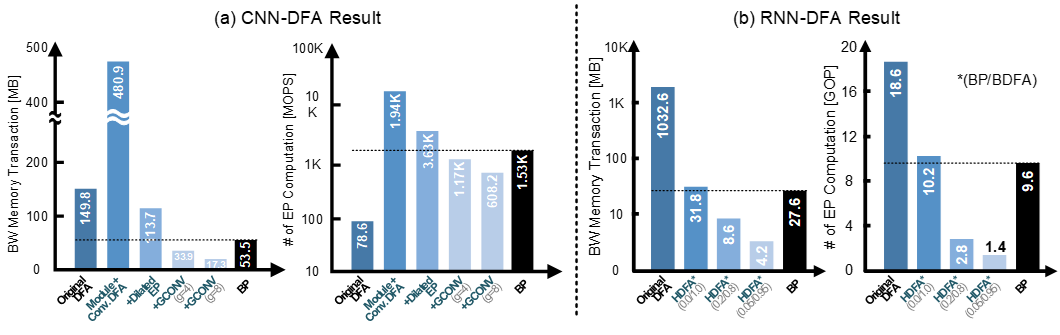}
\caption{Comparison of Hardware Efficiency (a) CNN-DFA: ResNet50, (b) RNN-DFA: GRU}
\label{fig_rnn_hw_spec}
\end{figure}  

\begin{figure}[t]
\includegraphics[width=\textwidth]{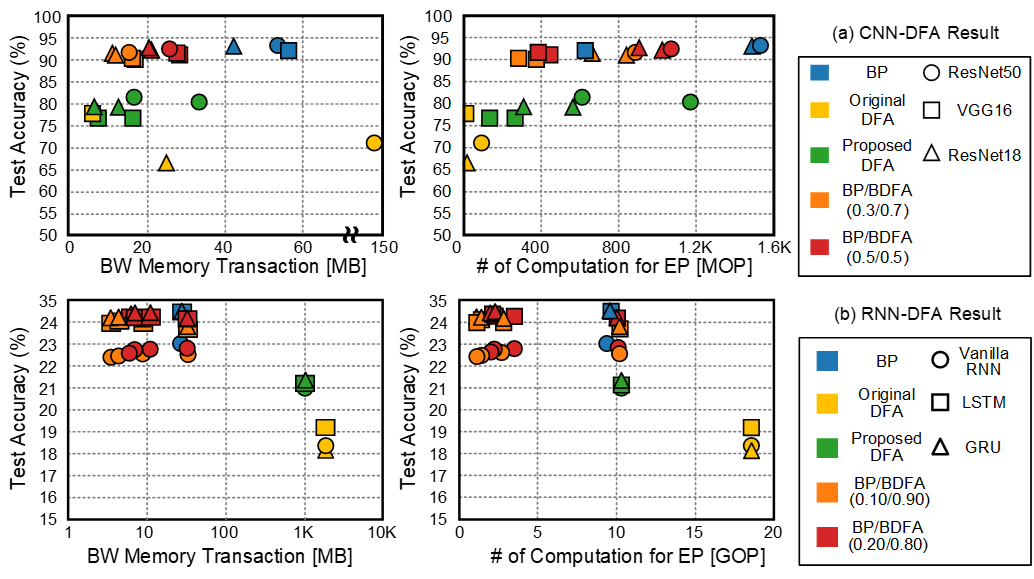}
\caption{Performance Summary of the Proposed Algorithm: (a) CNN-DFA, (b) RNN-DFA}
\label{fig_rnn_summary}
\end{figure}

\section{Conclusion}
\label{conclusion}
In this work, we develop the new DFA algorithm to be applied to the two general DNNs: CNN and RNN. At first, we divide the network into small fractional modules and connect the direct error propagation path only inside the module. We apply the dilated convolution and channel division to optimize the error propagation of the CNN. In the RNN training scenario, we disconnect unnecessary connections which disturb the accurate training and minimize the backward weight complexity for the low hardware cost. Finally, we suggest HDFA, which mixes both BP and DFA based training method in order to achieve high accuracy as same as the conventional BP. Thanks to the synergy of the hybrid training method, our proposed training method is biologically plausible and also achieve BP-level training performance with the lower hardware cost. The proposed algorithm has the potential of energy-efficient DNN learning acceleration with the application-specific integrated circuits such as \cite{DF-LNPU}. In addition, our proposed method solved the backward locking problem that appeared during the error propagation stage, so the acceleration of error propagation can be possible with the multi-GPU processing. 

In this research, we focused on the training of CNN and RNN from scratch. We will continue the research about the potential of the DFA algorithm in the distributed learning scenarios. For example, fine-tuning or federated learning (\cite{federated_learning}) can be examples of our application. Since we verified that the DFA algorithm shows lower hardware cost compared with the BP algorithm, we will develop the DFA algorithm for the light fine-tuning or federated learning methodology.

\subsection*{Acknowledgement} This work was supported by the Samsung Research Funding \& Incubation Center for Future Technology under Grant SRFC-TB1703-09.

\bibliographystyle{unsrtnat}
\bibliography{references}

\newpage
\begin{appendices}
\section{Direct Feedback Alignment based Convolutional Neural Network Training}
\subsection{Synergy of the DFA with BP}

Figure 7 shows the learning curve to demonstrate the effectiveness of HDFA. It is clear that using only DFA to train the network degrades the network performance critically. However, by combining the DFA with BP, this accuracy degradation can be compensated. To further verify that using BP and DFA together does not interrupt each other, we conducted an additional experiment. We design the BP only training scenario which assumes the weight gradient of the DFA is ignored to update the new weight. In this scenario, BP is occasionally performed to update the network, and weight is not updated when  DFA is applied. As shown in figure 7, the HDFA shows a faster and higher learning curve compared with the BP-only cases. It concludes that the HDFA can achieve accuracy comparable with BP thanks to the synergy of the two different learning algorithms.

\begin{figure}[hbt!]
\includegraphics[width=\textwidth]{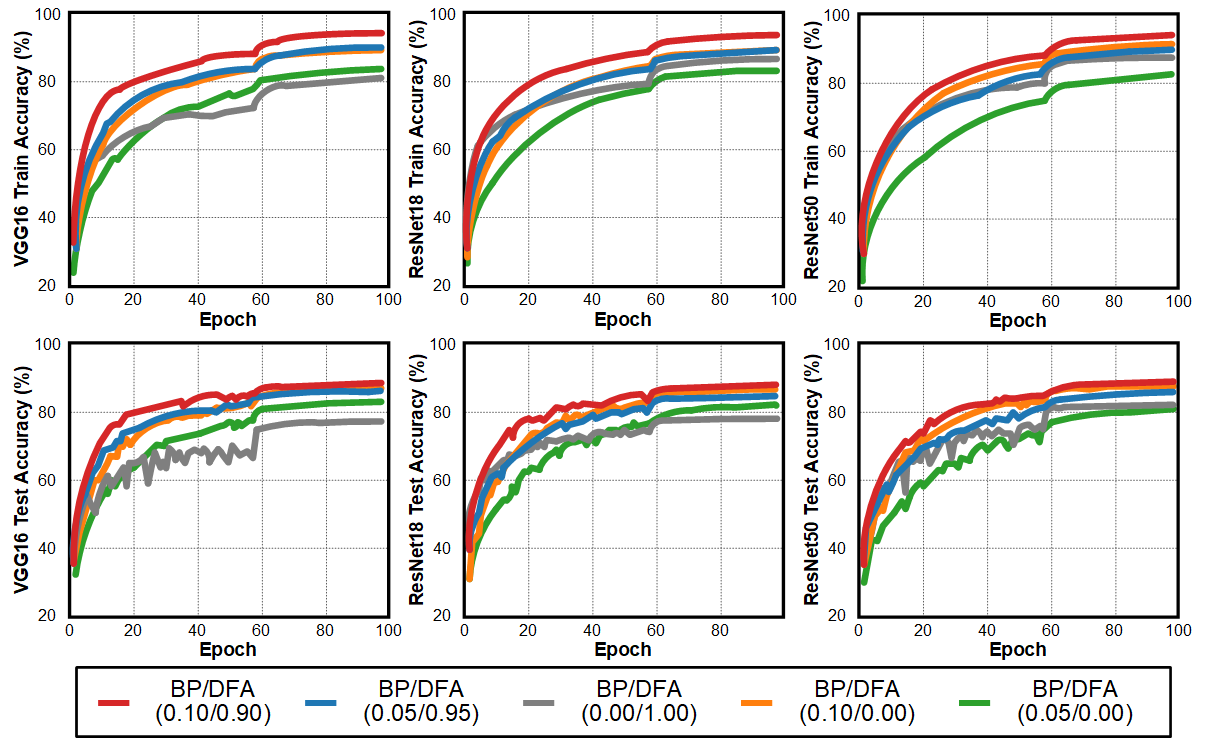}
\caption{CNN Training and Test Accuracy Curve with the Proposed HDFA}
\label{fig_cnn_sup1}
\end{figure}

\subsection{Scalability issue of the DFA}

DFA proposed by \cite{DFA} suffers from the feedback weight scalability issue on the CNN network. From figure 8 which compares the feedback weight ratio between the original DFA and our proposed DFA, two observations can be found. First, the proposed DFA is more memory efficient in convolution layers at the front. Second, proposed DFA becomes more beneficial as the CNN network becomes deeper. Motivated by this finding, a detailed analysis of feedback weight scalability is studied as follows. 

Denote \{width, height, channel\} of output error $\ve_{i}$ as \{$w, h, ich$\}, kernel size as $k$, number of channels at the input error and number of output classes as $och$, $class$ respectively. The number of feedback weights for the original DFA is expressed as $class \times w \times h \times ich$. The number of feedback weight for DFA introduced in Section 3.2 is $k \times k \times och \times ich \times 1/G$ where G is the number of groups used in channel division. As a result, the ratio of feedback weight $\bm{D}$ used in two different DFA methods is determined by the below equation.
    \begin{equation}
    ratio(\bm{D}^{DFA}_{ori}/\bm{D}^{DFA}_{proposed}) = (w \times h) / (k \times k) \times (class / och)
    \end{equation}
Figure 9 implies that the original DFA becomes inefficient as the input feature size and the number of the output class scales due to equation (13).

\begin{figure}[hbt!]
\includegraphics[width=\textwidth]{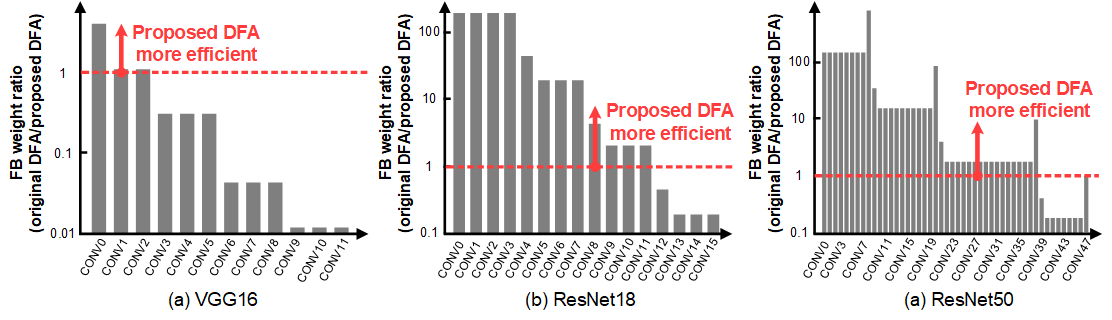}
\caption{Comparison of Feedback Weights according to Two Different DFA Methods}
\label{fig_cnn_sup2}
\end{figure}

\begin{figure}[hbt!]
\includegraphics[width=\textwidth]{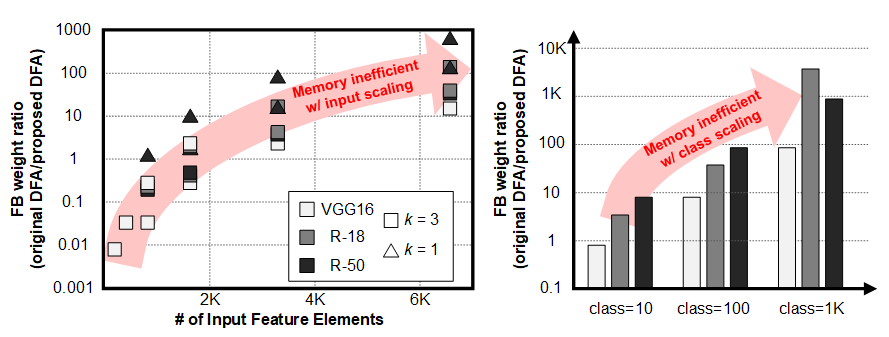}
\caption{DFA Scaling Issue with Two Different Factors : Input Feature Size (Left), Number of Classes (Right)}
\label{fig_cnn_sup3}
\end{figure}

\section{Direct Feedback Alignment based Recurrent Neural Network Training}
\subsection{Synergy of the DFA with BP}
Since the DFA has a different backward weight compared with the BP, it can be the obstacle of the training. We did the experiment to prove that the DFA and BP create a synergy effect. Figure 10 shows the results of the experiment. In this experiment, we compare the training performance of the HDFA and the BP only training scenarios. In the HDFA scenario, only a certain portion of iterations are trained with the BP algorithm, but the other portion is done by DFA. In contrast, the BP only training scenario performs only the BP based weight update. As shown in Figure 10, HDFA show faster and higher training curve compared with the BP is only performed. Furthermore, HDFA shows even higher performance compared with the DFA only training scenario. This curve can prove that the combination of the DFA and BP algorithms can achieve BP-level training accuracy with the low hardware cost. 
 
\begin{figure}[hbt!]
\includegraphics[width=\textwidth]{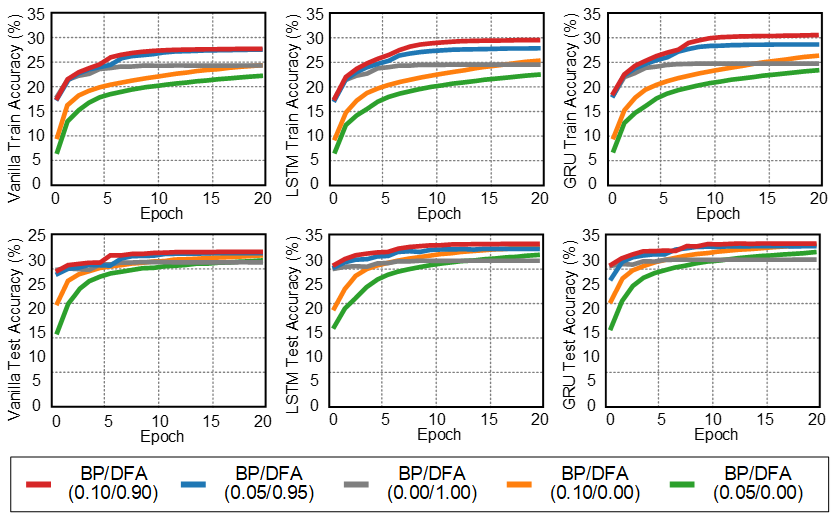}
\caption{RNN Training and Test Accuracy Curve with the Proposed HDFA}
\label{fig_rnn_curve}
\end{figure}

\subsection{Backward Weight Initialization Method}
\cite{BDFA} proposed a backward weight initialization method which is only applicable for fully-connected layers. It determines backward weight by assuming that all non-linear functions are removed. We also adopted the backward weight initialization concept for RNN training. Since we divide the last decoding fully-connected layers from the direct error propagation path, the backward weight used in the last layer is initialized with the transposed form of its forward weight. Of course, the backward weight is determined at the start of the training and maintains its values until training is finished. In the LSTM and GRU training, there are multiple forward weights in the RNN cell, so we initialize the backward weight by using the average value of the multiple forward weights. The final performance improvement details which include the backward weight initialization is depicted in Figure 11.

\begin{figure}[hbt!]
\includegraphics[width=\textwidth]{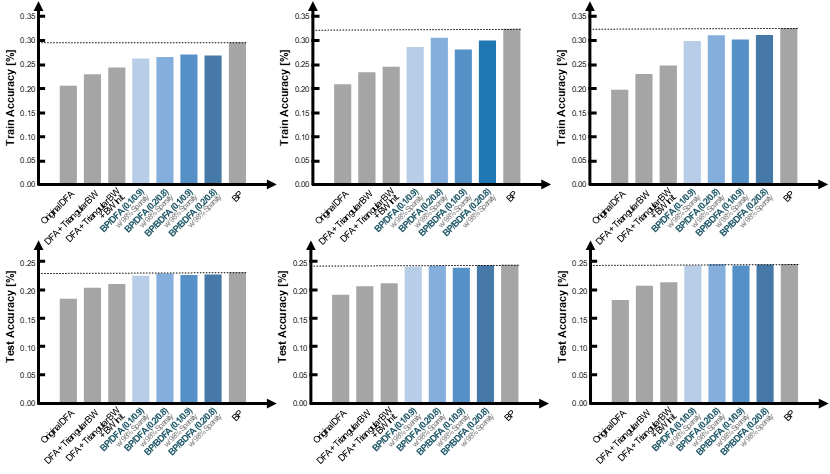}
\caption{Performance Improvement Details of DFA based RNN Training}
\label{fig_rnn_sup1}
\end{figure}

\subsection{Effect of Backward Weight Sparsity}
Figure 12 shows the training and test accuracy according to various backward weight sparsity. The effect of the sparsity becomes influential when the portion of BP is small. Additionally, vanilla RNN is vulnerable when the large sparsity appears in the backward weight. Another observation is that backward weight sparsity affects training accuracy but not the affects the test accuracy. 

\begin{figure}[hbt!]
\includegraphics[width=\textwidth]{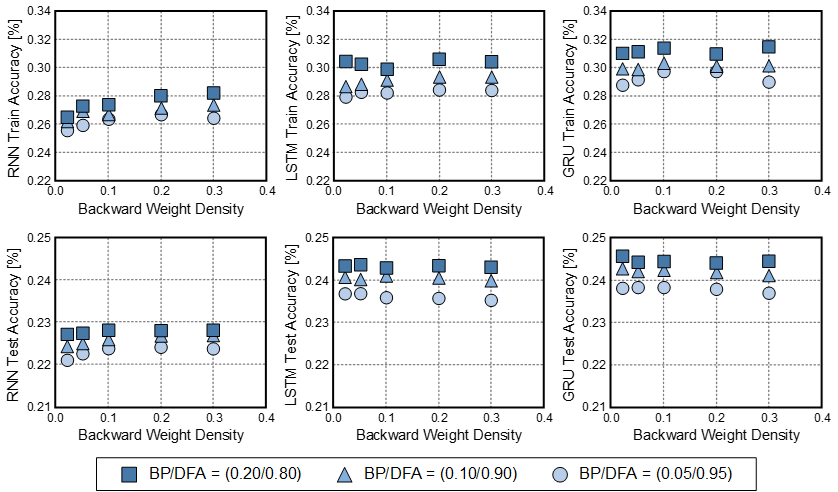}
\caption{Training \& Test Accuracy of RNN according to Backward Weight Sparsity \& BP/DFA Ratio}
\label{fig_rnn_sup2}
\end{figure}

\end{appendices}
\end{document}